\documentclass[letterpaper, 10 pt, conference]{ieeeconf}  % Comment this line out draftcls
\IEEEoverridecommandlockouts                              % This command is only
\usepackage[dvips]{graphicx}
\usepackage{verbatim}

\usepackage{xcolor}

\usepackage{amsmath, amssymb}
\usepackage{bm} %\bm command bold
\usepackage{hyperref}

\usepackage{dblfloatfix}    % To enable figures at the bottom of page

\usepackage{graphics} % for pdf, bitmapped graphics files
\usepackage{epsfig} % for postscript graphics files

\usepackage{algpseudocode}
\usepackage{algorithm}

\definecolor{my_gray}{gray}{0.8}

\usepackage{mathtools}

\def\yen{{\setbox0=\hbox{Y}Y\kern-.97\wd0\vbox{\hrule height.1ex
			width.98\wd0\kern.33ex\hrule height.1ex width.98\wd0\kern.45ex}}}

\title{\LARGE \bf
Blending Primitive Policies in Shared Control\\ for Assisted Teleoperation
}

\author{Guilherme Maeda$^{1*}$
\thanks{ $^1$Preferred Networks Inc. 1-6-1 Otemachi, Chiyoda, Tokyo, Japan. {\tt\footnotesize  gjmaeda@preferred.jp}. $^*$Corresponding author. 
}
}

\begin{document}

\maketitle
 \thispagestyle{empty}
\pagestyle{empty}

\begin{abstract}
	Movement primitives have the property to accommodate changes in the robot state while maintaining attraction to the original policy. As such, we investigate the use of primitives as a blending mechanism by considering that state deviations from the original policy are caused by user inputs. As the primitive recovers from the user input, it implicitly blends human and robot policies without requiring their weightings---referred to as arbitration. In this paper, we adopt Dynamical Movement Primitives (DMPs), which allow us to avoid the need for multiple demonstrations, and are fast enough to enable numerous instantiations, one for each hypothesis of the human intent. User studies are presented on assisted teleoperation tasks of reaching multiple goals and dynamic obstacle avoidance. Comparable performance to conventional teleoperation was achieved while significantly decreasing human intervention, often by more than 60\%.
\end{abstract}

\color{black}

\section{Introduction}

Recent events such as the Fukushima nuclear disaster and the COVID-19 pandemic have exacerbated the urgency for shared control and assisted teleoperation. In critical events, robot operators have to make decisions using incomplete and delayed information while operating robots through complex interfaces during long periods.
Direct teleoperation in such conditions is difficult if not impractical.
For years, the pursuit to make robots not only easier to remotely operate but also to decrease their dependence on the human operator has motivated research in shared control.
One of the fundamental challenges of shared control concerns how to algorithmically combine or ``blend'' the intentions of the human with the autonomous robot action.

\begin{figure}
	\centering
	\includegraphics[width=.85\linewidth]{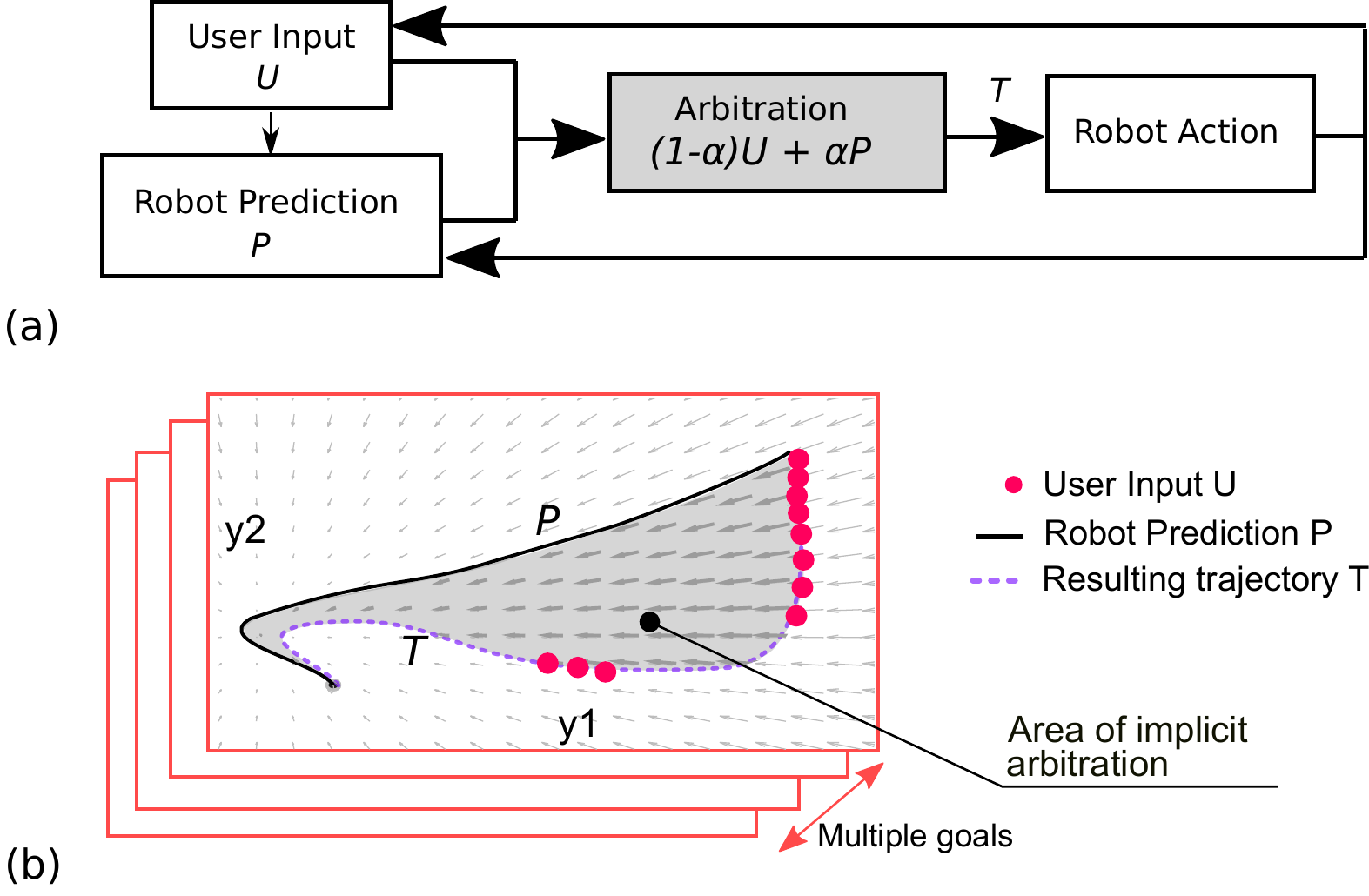}
	\caption{
		(a) Policy blending with an arbitration weight $\alpha$ as in \cite{draganPolicyblendingFormalismShared2013}.
		(b) A DMP vector field used to implicitly solve arbitration without the $\alpha$ parameter.
		The blending occurs as the robot recovers from a change of states induced by teleoperator commands.
		Multiple goals can be easily addressed by running independent DMPs with different goals but computed on the same robot's current state (indicated by the multiple layers).
	}	
	\label{fig:conceptarbitration}
\end{figure}

The problem of blending actions arises in a variety of forms. 
One form is when the robot acts autonomously in principle while allowing the user to intervene when necessary. 
Another is when the human is the main source of actions and the robot attempts to facilitate the operation. 
We introduce the blending problem with the conceptual framework of policy blending in \cite{draganPolicyblendingFormalismShared2013} (also shown in Fig. \ref{fig:conceptarbitration}(a)):
\begin{equation}
	(1-\alpha)U + \alpha P = T
		\label{eq:anca}
\end{equation}
where the arbitration weight $\alpha$ is used to combine the human input $U$ with the robot's predicted policy $P$ into a final policy $T$.

One should not be deceived by the simplicity of the policy blending formalism. 
Computing $\alpha$ is usually the challenging part because the arbitration is influenced by the context of the task, the skills, the intention and the preferences of the user, and the uncertainty of the robot's policy, to cite a few.

\subsection{Blending and Arbitration}

Devising an explicit arbitration function involves ingenuity and understanding of the task at hand. 
For example, Dragan et al. \cite{draganPolicyblendingFormalismShared2013} proposed arbitration as a function of the confidence of the robot, which increases as it approaches the inferred goal. 
On the other hand, in \cite{jainProbabilisticHumanIntent2019}, arbitration was computed as the confidence of the robot by comparing the probability of multiple goals.
%In both cases, the robot becomes more confident as the goal is approached, thus shifting the control towards itself and giving less control to the operator. 
When relying on domain knowledge, a hard-coded approach is to distribute the multiple degrees-of-freedom of a task between the human and the robot by assigning different arbitration values to each motion direction \cite{pervezMotionEncodingAsynchronous2019}.

Arbitration can be analytically derived from optimal control as it was done in \cite{liContinuousRoleAdaptation2015a} to minimize the interaction forces between the robot and the operator,
allowing the robot to switch roles from follower to leader depending on the amount of disagreement with the human.
Similarly, sharing effort strategies are closely related to arbitration in the force space. 
Those methods are characterized by adapting the impedance at the interaction interface, which can be based on the forces measured directly at the point of contact \cite{mortlRoleRolesPhysical2012} or by measurements of muscle activation via EMG \cite{peternelMultimodalIntentionInterfaces2016}, for example.

A simple and time-consuming approach is to brute-force search the optimal value of $\alpha$ as in \cite{xiRoboticSharedControl2019} and fix the optimized arbitration value throughout the task.
While being less stringent on training data, such a task-dependent approach suffers from generalization. 
Our experiments described in Section \ref{sec:toyproblem} show that transferring arbitration values even in simple contexts is problematic. % which also motivates our method.

A large part of the literature in shared control uses data from previous experiences to learn arbitration either directly or indirectly.
Arbitration can be related to the uncertainty of the intended goal assessed from a posterior distribution conditioned on the operator action \cite{draganPolicyblendingFormalismShared2013, Maeda2017AURO}.
Probabilistic models can be used to regulate the level of arbitration based on the deviation of the operator given a distribution of previously observed demonstrations \cite{farrajLearningbasedSharedControl2017} and also on the confidence of the robot on the task \cite{owanUncertaintybasedArbitrationHumanmachine2015}.
Past data can also be used to directly learn the arbitration function itself \cite{ohLearningArbitrationShared2019}.

Shared control also involves methods that do not blend policies.
Oh et al. \cite{oh2020natural} referred to such methods as ``indirect blending''.
Here, the robot prediction/policy is computed as a function of the human action, as a sequential process as opposed to the simultaneous nature of blending.
In this category, Javdani et al.~\cite{javdaniSharedAutonomyHindsight2015} framed assistance as a POMDP where the intended operator goal is an unobserved state.
Policies that augment the human action have been proposed in \cite{oh2020natural, schaffResidualPolicyLearning2020}.
Other approaches include directly generating a robot trajectory with primitves \cite{yangOnlineAdaptiveTeleoperation2019} or by motion planning \cite{youAssistedTeleoperationStrategies2012} by first predicting the user intention.

\subsection{Shared Control using Primitives}

Recent approaches to virtual guides and virtual fixtures (e.g. \cite{aarnoAdaptiveVirtualFixtures2005, zeestratenProgrammingDemonstrationShared2018, ewertonAssistedTeleoperationChanging2020}), and trajectory-level task representations (e.g. \cite{farrajLearningbasedSharedControl2017, hauserRecognitionPredictionPlanning2013, pervezMotionEncodingAsynchronous2019})
usually rely on demonstration of trajectories to build models that capture the primitive patterns within the demonstrations.
Thus, by assuming that the teleoperator preferred behavior is to be within the distribution of previous demonstrations, virtual guides attempt to constrain the teleoperator commands usually by adjusting the impedance on the interface \cite{ewertonAssistedTeleoperationChanging2020}.
Offline, this assumption also allows a system to discard new data or incrementally enrich the  current primitive distribution of trajectories \cite{farrajLearningbasedSharedControl2017}.

Although the use of demonstrations is certainly useful in assisted teleoperation, the dependence on user demonstrations is also one of the main hurdles to deploying data-driven teleoperation systems in practice. Data-driven methods that rely on multiple demonstrations---often recorded by teleoperation itself---carry the assumption that an underlying behavior exists while rendering learned models task-dependent. In real cases, urgent deployment in scenarios such as those found in disaster response may not provide conditions for demonstrations in the first place.

In this work, we adopt Dynamical Movement Primitives (DMPs) \cite{ijspeert2013dynamical} due to its widespread use and ease of implementation.
As shown in Fig. \ref{fig:conceptarbitration}(b), a DMP can gradually assimilate a disturbed state towards the original path due to its characteristic vector force field.
If the disturbed state is a function of $U$, the recovering of the primitive towards the original robot policy $P$ implicitly blends the user input command  without computing arbitration values.
By comparing the two concepts (a) and (b) in Fig. \ref{fig:conceptarbitration} we can interpret the automatic arbitration provided by a DMP as the net result of the interacting forces between the force field and the intensity of the user input command (the gray area in subplot (b)).

\subsection{Contribution}

\emph{Policy Blending with Primitives (PBP)} is a scalable, online, and simple method that implicitly solves arbitration in shared control.
Our technical contribution is an algorithm that exploits the disturbance rejection properties of primitives as a policy blending mechanism.
By using DMPs as a base, the method both frees the designer from designing \emph{task-specific arbitration functions} and from \emph{the need to provide multiple demonstrations} while addressing \emph{scenarios with multiple goals}.
Ultimately, the relaxation on arbitration and demonstration requirements aims at bringing assisted teleoperation closer to real deployment scenarios, particularly for cases in which prior training of task-dependent models is not practical. 
Our results show that PBP is effective even on tasks for which demonstrations are unlikely to be consistent among users.

\section{Policy Blending with Primitives}

For each possible goal in a task PBP creates a Dynamical Movement Primitive (DMP) as

\begin{equation}
	\tau \ddot y = K_p (g - y(\cdot, u)) - K_d \dot y(\cdot, u) + f,
	\label{eq:dmp}
\end{equation}
where $\tau$ is a time constant that governs the speed of execution, $K_p$ and $K_d$ are positive scalar values that modulate a linear damped attraction towards the goal $g$. The forcing function $f$ adds nonlinearities to the robot movement.
The nonlinear behavior is usually acquired by demonstrations by isolating $f$ to the left side and replacing the values of $y$, $z$, and $g$ with those from the demonstration (see \cite{ijspeert2013dynamical} for a thorough overview and explanation of the method). 
The notation $y=y(\cdot, u)$ is to indicate that in PBP, the robot state is  a function of the operator command $u$ and two variations will be discussed on how this can be accomplished.

\subsection{Arbitration in PBP}

The arbitration analysis of a PBP is non-trivial because of the existence of a nonlinear function $f$ and because the operator command only indirectly affects the robot state via $y(\cdot, u)$.
However, for the particular case when the DMP reaches the steady-state and regulates the robot on the goal attractor $g$, the forcing function $f$ goes to zero by construction~\cite{ijspeert2013dynamical} and Eq. \eqref{eq:dmp} reduces to a proportional feedback law
\begin{equation}
	K_p(g-y(\cdot,u))=0 \label{eq:ss}.
\end{equation}

Defining $K_p = (1-\alpha)$ the feedback law becomes
\begin{equation}
	(1-\alpha)y(\cdot,u)-\alpha g = g. 
	\label{eq:p}
\end{equation}
By comparing Eqs. \eqref{eq:anca} and \eqref{eq:p} we see that the steady-state DMP is similar to the original blending concept.
If $T\!=\!P$ in Eq. \eqref{eq:anca}, then the original policy blending can be written as \mbox{$K_p(P-U)\!=\!0$}, as a feedback controller that  minimizes the difference between the robot prediction and user input as in the PBP case.

This comparison shows that PBP automatic arbitration is due to the feedback nature of a DMP. 
The feedback also imposes a goal-directed behavior where the final state of the robot policy must be the goal itself, $P(t\rightarrow\infty)=g$.
The policy blending of Eq. \eqref{eq:anca} allows for the blended output $T$ to be an arbitrary state, such as a time-varying goal state, at the expense that $\alpha$ becomes an open parameter.

\subsection{Continuous and Alternated Blending}

We investigated two possible ways to realize  $y = y(u)$ in practice.
Algorithm 1 shows an elementary loop where at each time step the primitive $P$ computes a new state towards the goal by integrating Eq.  \eqref{eq:dmp}.
Within the same control cycle, user commands $u$ are read from the teleoperation interface. The actual state obtained from the robot low-level controller results from tracking a reference signal that is a continuous combination of both DMP state $y$ and the user input $u$.

In its simplest form, $y$ and $u$ both belong to the same space of positions. 
The DMP predicts the next robot recovering position $y$, and the user disturbs the robot by moving it to a new position $u$.
The function $ROBOT.control(y + u)$ is a reference tracker where the reference is $y+u$.

\begin{algorithm}[H]
{\fontsize{9pt}{9pt}\selectfont	
	\caption{CONTINUOUS\_BLENDING} \label{alg:continuous}
	\begin{algorithmic}[1]
		\State $y'    \leftarrow  ROBOT.initial\_state()$
		\While 1:
			\State $y    \leftarrow   P.next\_step(g, y')$ \color{gray}(note that $y=y(\cdot, u)$) \color{black}
			\State $u    \leftarrow   USER.read\_input()$	
			\State $y' \leftarrow  ROBOT.control(y + u)$
		\EndWhile		
	\end{algorithmic}
}
\end{algorithm}

Algorithm 2 implements a variation in which the primitive $P$ only proceeds if there is no human input. In the case the user intervenes, the robot ``detaches'' from the primitive and locks its phase, allowing the operator to freely move the robot as if it was in full teleoperation mode such that $y=u$.
Once the operator releases the controller the primitive resumes based on the new state, implicitly blending the disturbed states into the original policy $P$ until the goal $g$ is achieved such that $y=y(u)$.
\begin{algorithm}
	{\fontsize{9pt}{9pt}\selectfont		
		\caption{ALTERNATED\_BLENDING} \label{alg:alternated}
		\begin{algorithmic}[1]
			\State $y'    \leftarrow  ROBOT.initial\_state()$
			\While 1:
			\State $u    \leftarrow   USER.read\_input()$	
			\If{ $u$ == 0}
			\State $y    \leftarrow   P.next\_step(g, y')$ \color{gray}(note that $y=y(\cdot, u)$) \color{black}
			\State $y' \leftarrow  ROBOT.control(y)$
			\Else
			\State $y' \leftarrow  ROBOT.control(u)$ \color{gray}(note that $y \sim u$) \color{black}
			\EndIf
			\EndWhile
		\end{algorithmic}
	}
\end{algorithm}

\begin{figure}
	\centering
	\vspace{.3cm}
	\includegraphics[width=0.95\linewidth]{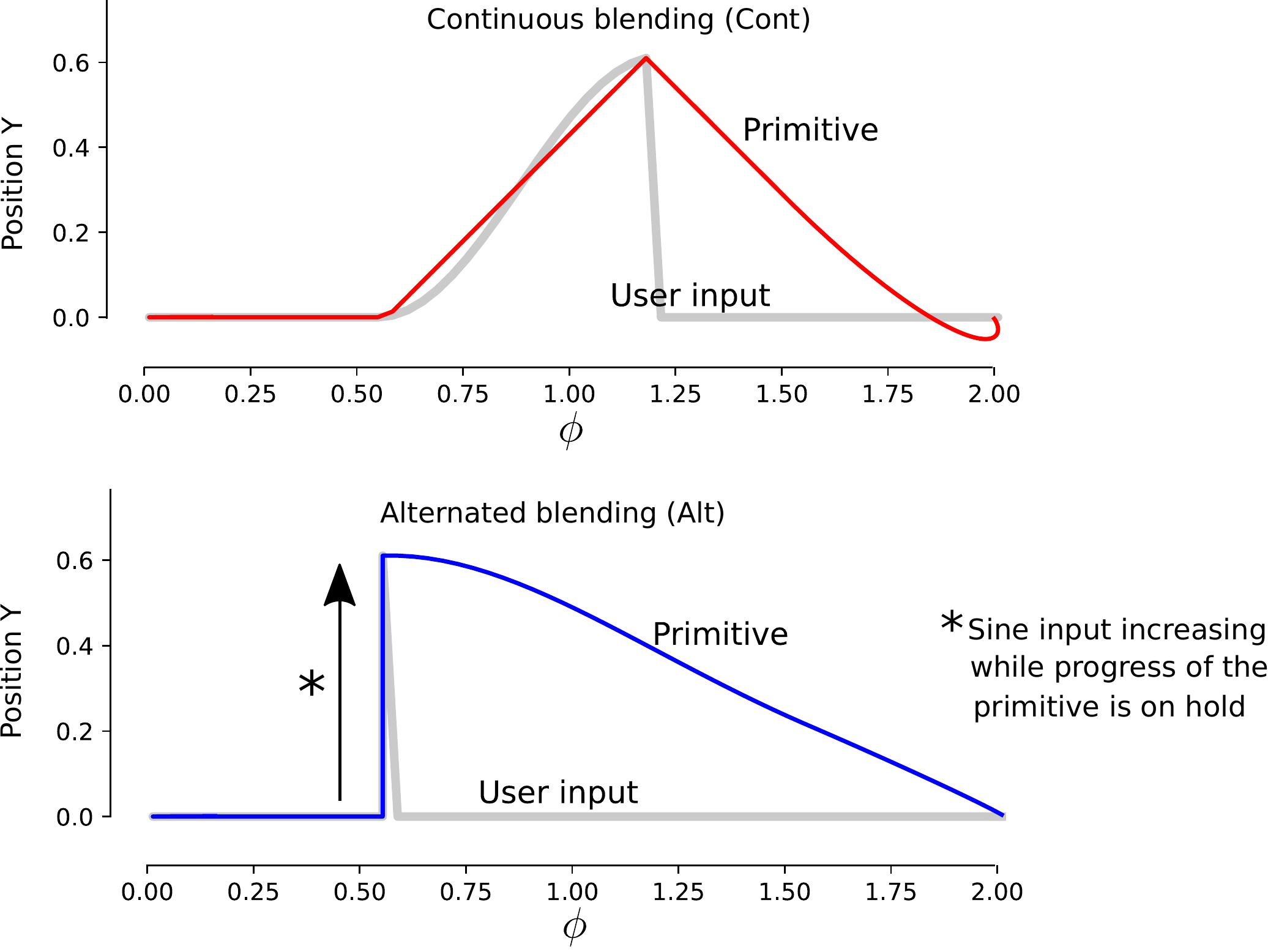}
	\vspace{-.3cm}
	\caption{
		The difference in behavior between the alternated and continuous blending methods. Both primitives are subject to the same user input as a time-varying open-loop sinusoidal step. 
	}
	\label{fig:disturbanceeffect}
\end{figure}

Fig. \ref{fig:disturbanceeffect} graphically shows the difference between the two methods.
In the continuous blending, the DMP progresses as the value of $U$ increases according to the phase $\phi$.
When the user input suddenly goes to zero at around  $\phi\!=\!1.25$ the DMP blends the residual external effects by converging to the final goal attractor.
In the alternated blending, the DMP locks its phase while the operator acts in the Y direction and restarts blending once the user input vanishes.
The resulting paths are strikingly different despite the same user inputs.

\subsection{Multiple Goals}

Tracking multiple goals is essential in assisted teleoperation not only because the human intention is not known to the robot but also because the intended goal may change during the task.
To switch among primitives without discontinuities we allocate one primitive to each candidate goal---illustrated in Fig. \ref{fig:conceptarbitration}(b).
At each point in time, the robot executes the primitive whose goal is the most likely one.
Algorithm \ref{alg:continuous_multi_goal} presents a version with $K$ possible goals for the continuous blending case (a similar modification applies to the alternated blending case).
\begin{algorithm}
{\fontsize{9pt}{9pt}\selectfont		
	\caption{MULTIPLE\_CONTINUOUS\_BLENDING} \label{alg:continuous_multi_goal}
	\begin{algorithmic}[1]
		\State $y'    \leftarrow  ROBOT.initial\_state()$
		\While 1:
		\For {$k=1$ to $K$} 
		\State $y_k   \leftarrow   P_k.next\_step(g_k, y')$	
		\EndFor
		\State $u    \leftarrow   USER.read\_input()$	
		
		\State $y^* \leftarrow GOAL\_ESTIMATOR(y_{\{1,..,K\}},g_{\{1,..,K\}},u) $
		
		\State $y' \leftarrow  ROBOT.control(y^* + u)$
		\EndWhile		
	\end{algorithmic}
}
\end{algorithm}

\begin{figure*}
	\centering
	\vspace{.3cm}
	\includegraphics[width=0.9\linewidth]{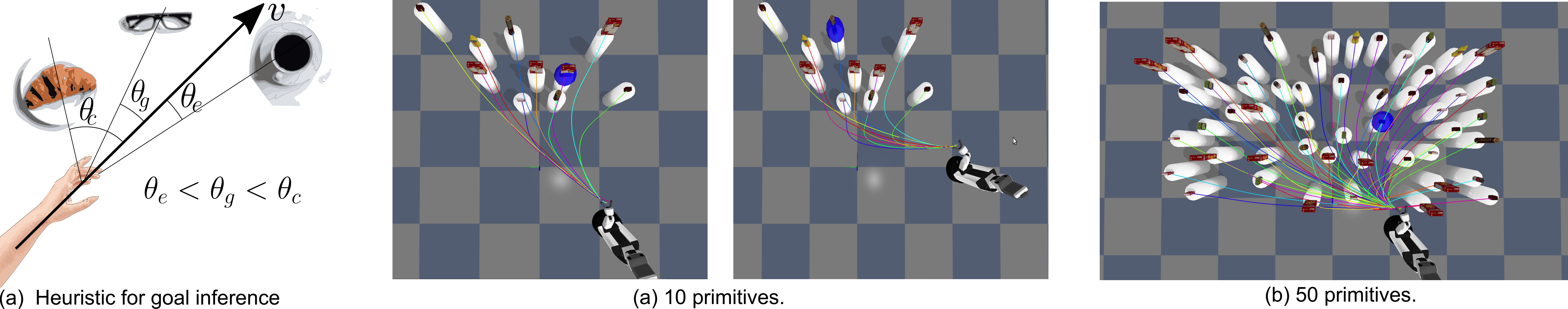}
	\vspace{-.25cm}
	\caption{
		(a) A simple heuristic is used to identify the desired object based on the angular distance to an object.
	    (b) Teleoperation under multiple possible goals. The blue ball indicates the estimated human goal. 
        (c) Primitives are computationally inexpensive allowing the tracking of 50 objects without changes in the control frequency.
	}
	\label{fig:multipledmps}
\end{figure*}

The function $GOAL\_ESTIMATOR$ determines which of the possible goals \mbox{$g^* \in \{g_1, .., g_K\}$}  is intended by the user, hence outputting the DMP state corresponding to that goal \mbox{$y^* \in \{y_1, .., y_K\}$}. 
For simplicity, here we assume that the direction in which the users point the robot's end-effector is an indicator of their intention\footnote{The motivation for proposing the described heuristic is mainly to realize the system in practice but other functions for goal estimation could also be used.}. 
This intuition is illustrated in Fig. \ref{fig:multipledmps}(a).
This reasoning can be implemented by computing the alignment angle between the robot end-effector's direction w.r.t. each of the tracked objects $\Theta \in \{\theta_1, ..., \theta_K\}$ at each time-step. 

Figure \ref{fig:multipledmps}(b) shows snapshots where 10 primitives are allocated to each object.
The blue ball indicates the object that is being dynamically selected by $GOAL\_ESTIMATOR$. 
Subplot (c) shows a snapshot where 50 objects are being tracked by their respective DMPs. 
We went as far as testing our method on 100 DMPs without observing any decrease in the loop frequency of 30 Hz of the controller, currently scripted in Python language.

\section{Experiments}

Using a simple low-dimensional problem, we illustrate the benefits of PBP concerning explicit arbitration.
Next, user studies are used to evaluate the method using a high-dimensional mobile manipulator under simulated dynamics.
We refer to PBP Alternated and Continuous implementations as PBP-Alt and PBP-Cont in the subsequent text. 
 
\subsection{Advantages over Explicit Arbitration}  \label{sec:toyproblem} 

\begin{figure}
	\centering
	\includegraphics[width=.95\linewidth]{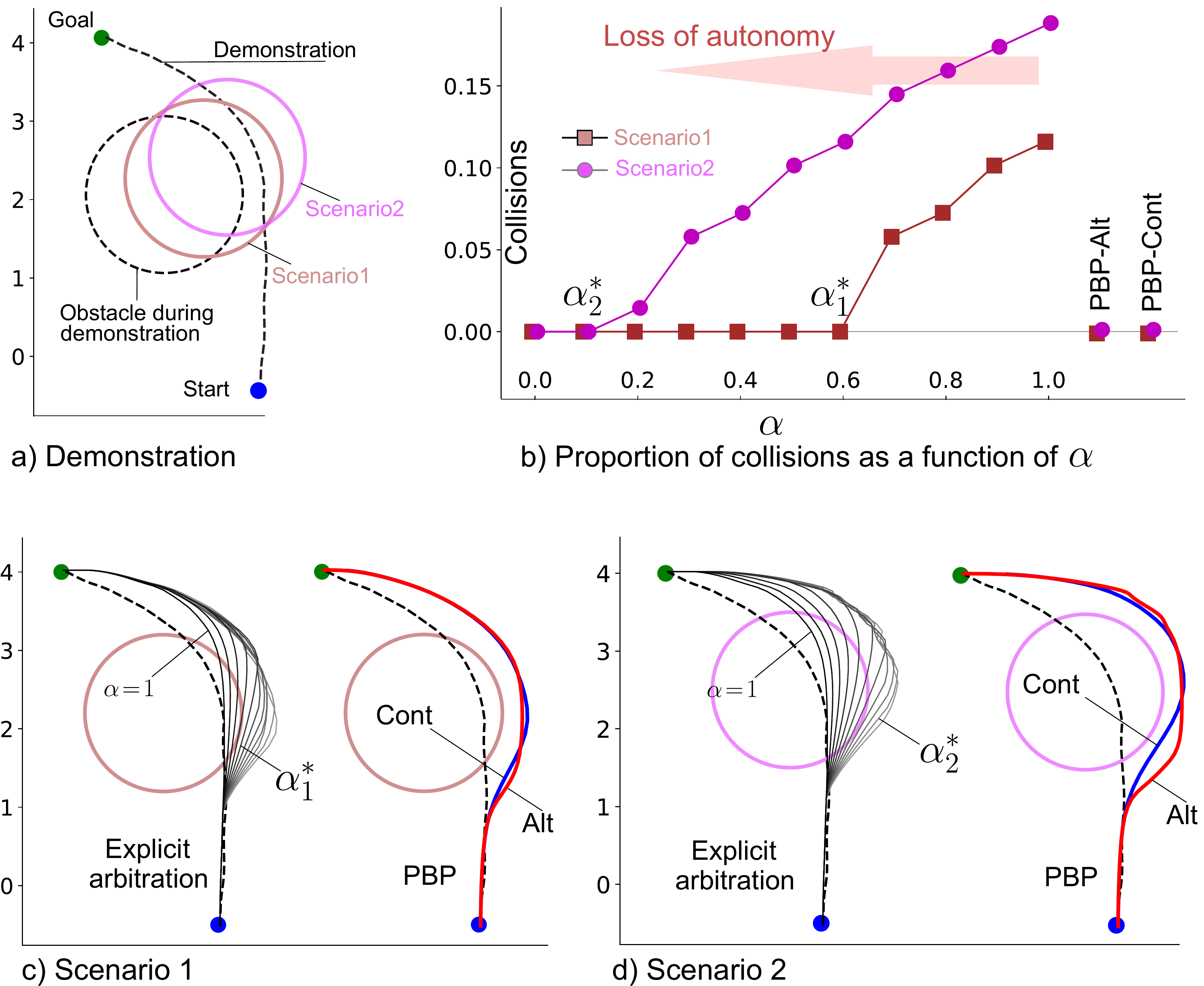}
	\vspace{-.04cm}
	\caption{
		(a) Demonstration scenario (dashed lines) and two obstacle placements under collision with the original demonstrated trajectory.
		(b) Using a virtual operator, the amount of collision decreases proportionally with the robot's loss of autonomy.
		(c-d) Trajectories under explicit arbitration and using the PBP method.
	}
	\label{fig:alphastudy}
\end{figure}

Figure \ref{fig:alphastudy}(a) shows a task where an operator demonstrated a path avoiding an obstacle given a start and goal states on an XY plane.
Two new scenarios were simulated, differing by the amount in which the obstacle was moved from its original position during test time.
By using the demonstrated trajectory as the robot policy $P$, collisions are expected in both scenarios in different amounts.

Using Eq.  \eqref{eq:anca}, we investigate the effect of $\alpha$ on the combined policy $T$ when $U$ is a command that attempts to avoid obstacles.
The motion of the robot is simulated by a mass-spring-damper and uses the sequences of states in the demonstration as a time-indexed open-loop reference input, providing the next state $P$ at each time step.
To deterministically simulate the operator input $U$, we created a virtual operator whose policy is 
\begin{align}
	U = \begin{dcases*}
		sign(\cdot) \times 1, & if $ d < 0.2 $,\\
		0, & otherwise. 
	\end{dcases*}
	\label{eq:virtual_human}
\end{align}
where the sign of the input moves the robot to the opposite direction from that of the obstacle and $d$ is the distance to the obstacle.

For each scenario, we ran Eq. \eqref{eq:anca} with $U$ given by Eq.  \eqref{eq:virtual_human} with  \mbox{$\alpha=\{0, 0.1, 0.2, ... 1.0\}$}.
When $\alpha=0$ the operator has total control when a collision is detected, and when $\alpha=1$ the operator input is ignored. Intermediate values blend the two policies in different ratios.
The ideal arbitration is the highest value of $\alpha$ that allows the virtual operator to avoid collisions since lower than necessary values decrease the autonomy of the robot.
As such, for Scenario 1, the ideal arbitration is $\alpha^*_1=0.1$ while for Scenario 2 the value is $\alpha^*_2=0.6$.
The result is summarized in Fig. \ref{fig:alphastudy}(b) as the ratio of collisions as a function of $\alpha$.

The results show that setting the correct $\alpha$ on even simple tasks is a non-trivial problem that is highly sensitive to variations in the scene.
On the other hand, when regressing $P$ as a DMP and using PBP (either alternated or continuous) with the same virtual human of Eq. \eqref{eq:virtual_human}, all resulting trajectories were collision-free (subplots (c) and (d)) without requiring the specification of an arbitration value.

\subsection{User Studies with a Mobile Manipulator}

PBP was investigated in a simulated environment with a Toyota HSR \cite{yamamoto2019development}.
PBP-Alt and PBP-Cont were compared against a teleoperated robot baseline (Teleop). 
Our study was conducted with 11 participants who did not have previous contact with the system. 
Five participants had a robotics background, five were researchers from non-robotics fields, and one was from humanities.
For the sake of simplicity and scalability, differently from the previous experiment where a demonstration was provided, all DMPs were initialized with straight trajectories connecting the current robot state to the goal state.

For each method, participants were allowed to operate the robot until they felt comfortable and confident with the operation. 
Once habituated, each participant operated the robot 15 times under randomly generated scenes using the same control method while trajectories and commands were recorded. Next, they switched to the other method, repeating the process of habituation and 15 trials. 
To eliminate biases due to the ordering of learning, each participant was presented with a different sequence of methods.
All participants went through the same scenarios by resetting the seed of the random generators to predefined values.

The user operated the robot from the robot's perspective.
The keyboard arrows were used to move the robot sideways and back and forth, and the mouse lateral motion was used to rotate the base of the robot. 
For the sake of fairness, when using the PBP-Alt and PBP-Cont, the forward key was disabled and the robot's forward motion was solely due to the progress of the primitive's phase. In Teleop, the forward key gain was calibrated such that the robot had the same speed as when the robot was driven by the DMPs in the PBP methods. Video of experiments can be watched in {\small \verb|https://youtu.be/rqq46_-Bwes|}.

\vspace{0.2cm}
\subsubsection{Reaching under Multiple Goals}
Here, the goal was to evaluate the ability of the operator to switch primitives to reach one out of multiple objects.
Each scene was generated by uniformly randomly placing two to four objects in front of the robot. Each object position was random sampled from an uniform distribution in the range $X= [1.25, 2.5]$ and $Y= [-1.05, 1.05]$ (in meters). The height was fixed to the same value for all objects. The target object was visually indicated to the operator by marking it with a yellow ball.

\begin{figure}
	\centering
	\vspace{.25cm}
	\includegraphics[width=0.999\linewidth]{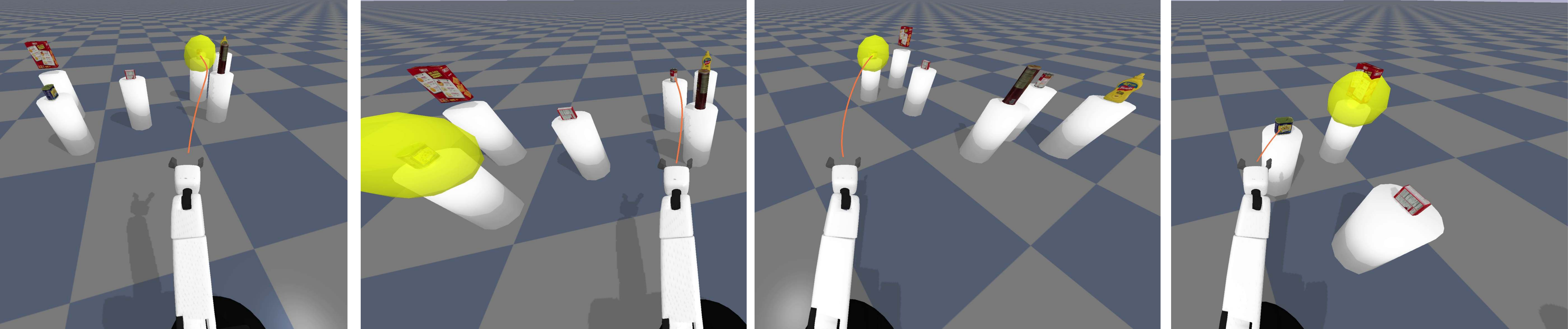}
	\vspace{-.6cm}
	\caption{
		A randomly generated scene with multiple goals. 		
		In the first snapshot, the path leads to the correct object marked by the yellow target ball. In the second snapshot, the target jumps to another object, forcing the user to intervene such that the robot selects the appropriate primitive in the third snapshot.		
	}
	\label{fig:grasptask}	
\end{figure}

We defined a progress schedule of $0\%, 30\%$, and $60\%$ towards the current goal. At each scheduled point, a new target object was randomly chosen, thus forcing the operator to switch primitives. The users were not informed about this schedule, so although they expected a possible change of the target, they did not know when or if it would happen.
The task finished when the operator could bring the end-effector within a distance of $5$ cm to the current target object.
Figure \ref{fig:grasptask} shows a typical instance of the task.
For the PBP-Alt and PBP-Cont methods, the DMP path of the currently estimated goal was plotted online to indicate to the operator the active primitive.

\begin{figure}
	\centering
	\vspace{.25cm}
	\includegraphics[width=0.999\linewidth]{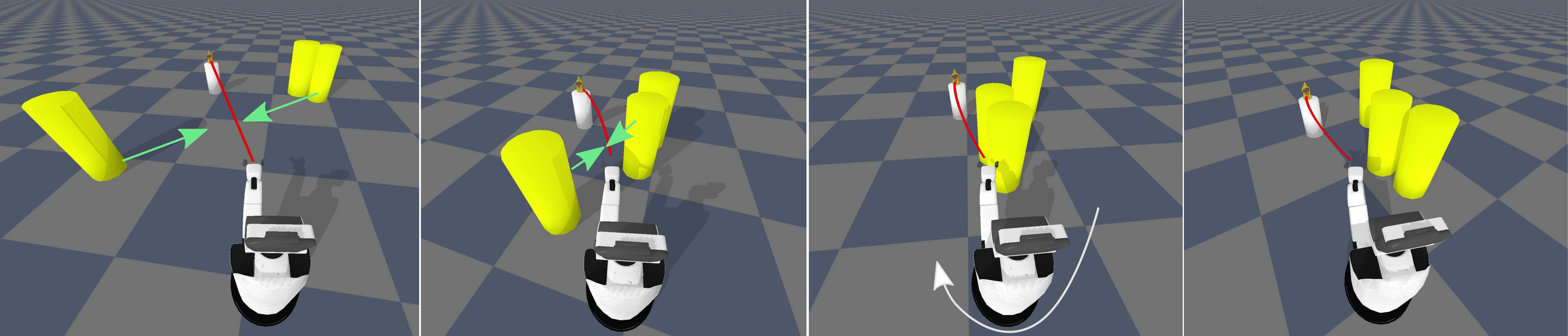}
	\vspace{-.5cm}
	\caption{
		In the obstacle task, the robot had to reach the object as the cylinder-shaped obstacles were randomly activated. 
		Once activated, the cylinders tried to intercept the robot, forcing the user to reshape the original DMP trajectory. 
	}
	\label{fig:obstacletask}
\end{figure}

\begin{figure*}[h]
	\centering
	\includegraphics[width=0.99\textwidth]{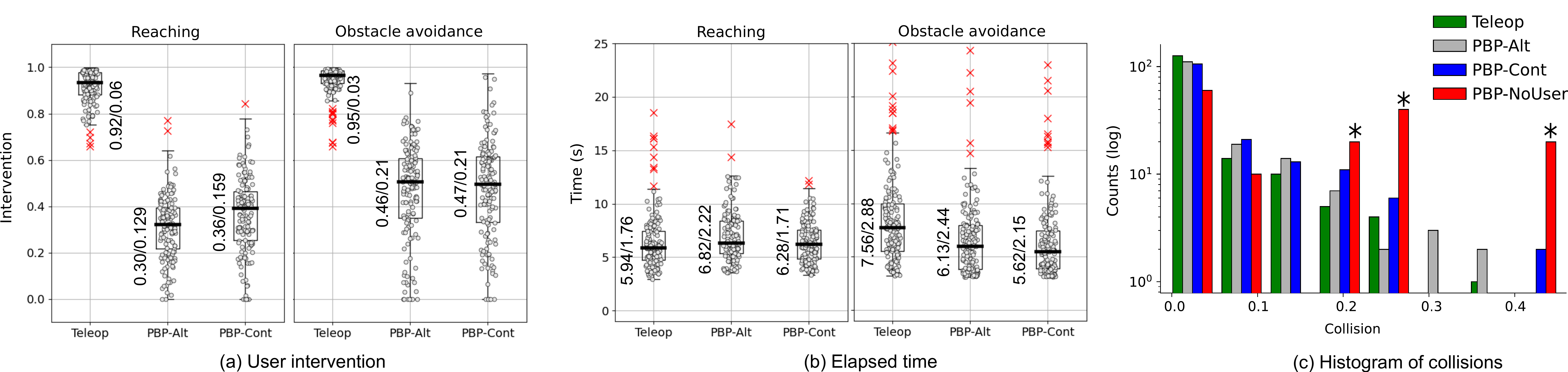}
	\vspace{-.2cm}
	\caption{
	(a-b) Summary of users' performance using Teleop, PBP-Alt, and PBP-Cont. The raw data
is shown as gray circles and outliers as red crosses. The vertical values close to each box show the
mean and one standard deviation of a normal approximation ($\mu/\sigma$).
The amount of user intervention during task (a) was greatly reduced in both reaching and obstacle avoidance tasks, without significant
differences in elapsed times (b).
	(c) The histogram of collisions during the obstacle avoidance task including a case where no user input was given (PBP-NoUser).
	}
	\label{fig:user_metrics_and_collision_ICRA}
\end{figure*}

\vspace{0.2cm}
\subsubsection{Obstacle Avoidance}
Here, the goal was to evaluate how easy it was for the operator to steer the robot and modify the primitive path $P$ to avoid dynamic obstacles.
While there are methods where DMPs have been used to autonomously avoid obstacles \cite{hoffmannBiologicallyinspiredDynamicalSystems2009}, automating obstacle avoidance would defeat the purpose of human intervention and thus shared teleoperation.
The task consisted of a single target object, positioned randomly at the beginning of each trial. 
Three randomly positioned obstacles started static but were randomly activated to move following a predefined robot's progress threshold. 
The obstacle motion intended to intercept the robot at the midpoint of a line connecting the current robot position and the goal.

The dynamics among the user commands, the robot policy, and the motion of obstacles make the resulting trajectory extremely hard, if not impossible to predict.
Users were forced to take multi-modal actions, at times pushing through obstacles at the risk of collision and at times backing off and waiting (see Fig. \ref{fig:obstacletask}).
This scenario makes it {unlikely to obtain a consistent set of demonstrations} that could be used, for example, as virtual guides or to build inference models.

\vspace{0.2cm}
\subsubsection{Results and Discussions on User Studies}

Figure \ref{fig:user_metrics_and_collision_ICRA} shows quantitative metrics on all users' data regarding (a) the amount of human intervention, and (b) the time it took them to finish the task. User intervention was computed as the ratio $U_{total}:T_{task}$, where $U_{total}$ is the total number of time steps in which the user was either using the keyboard or the mouse, and $T_{task}$ is the total number of time steps taken to accomplish the task. The values close to each box show the mean and one standard deviation ($\mu/\sigma$). As expected, under Teleop the ratios on user intervention are close to one ($\mu=0.92$ and $\mu=0.95$ for grasping and obstacle avoidance, respectively) but not exactly one due to periods in which the user was not taking any action. For the reaching task with the PBP methods, the user load was decreased by more than 60\%.
In comparison to reaching, the amount of interaction with PBP during obstacle avoidance increased by 20\% on average since users had to actively deviate from obstacles. 

Under the PBP methods, the task time did not increase as evidence that users could indicate to the robot the intended goal without loss of time. 
In the obstacle avoidance case, Teleop took on average about  1 to 2 seconds longer than the PBP methods. 
When pairing this result with the respective histogram of states under collision in Fig. \ref{fig:user_metrics_and_collision_ICRA}(c)
it is evident that although Teleop led to longer tasks, it also led to slightly fewer collisions.

To assess how users decreased collisions by sharing control with the primitives, we ran the same obstacle avoidance task using the PBP method where \mbox{$U\!=\!0$} at all times. The resulting histogram, labeled as ``PBP-NoUser'' in Fig. \ref{fig:user_metrics_and_collision_ICRA}(c), shows that without user intervention more than 50\%  of the trials presented at least 20\% of states under collision (the bins marked with an asterisk in the same Figure).  In contrast, the majority of trials with either PBP-Alt or Cont had no collision as indicated by the dominant magnitude of the leftmost bin.

\begin{figure}
	\centering
	\vspace{.03cm}
	\includegraphics[width=0.9\linewidth]{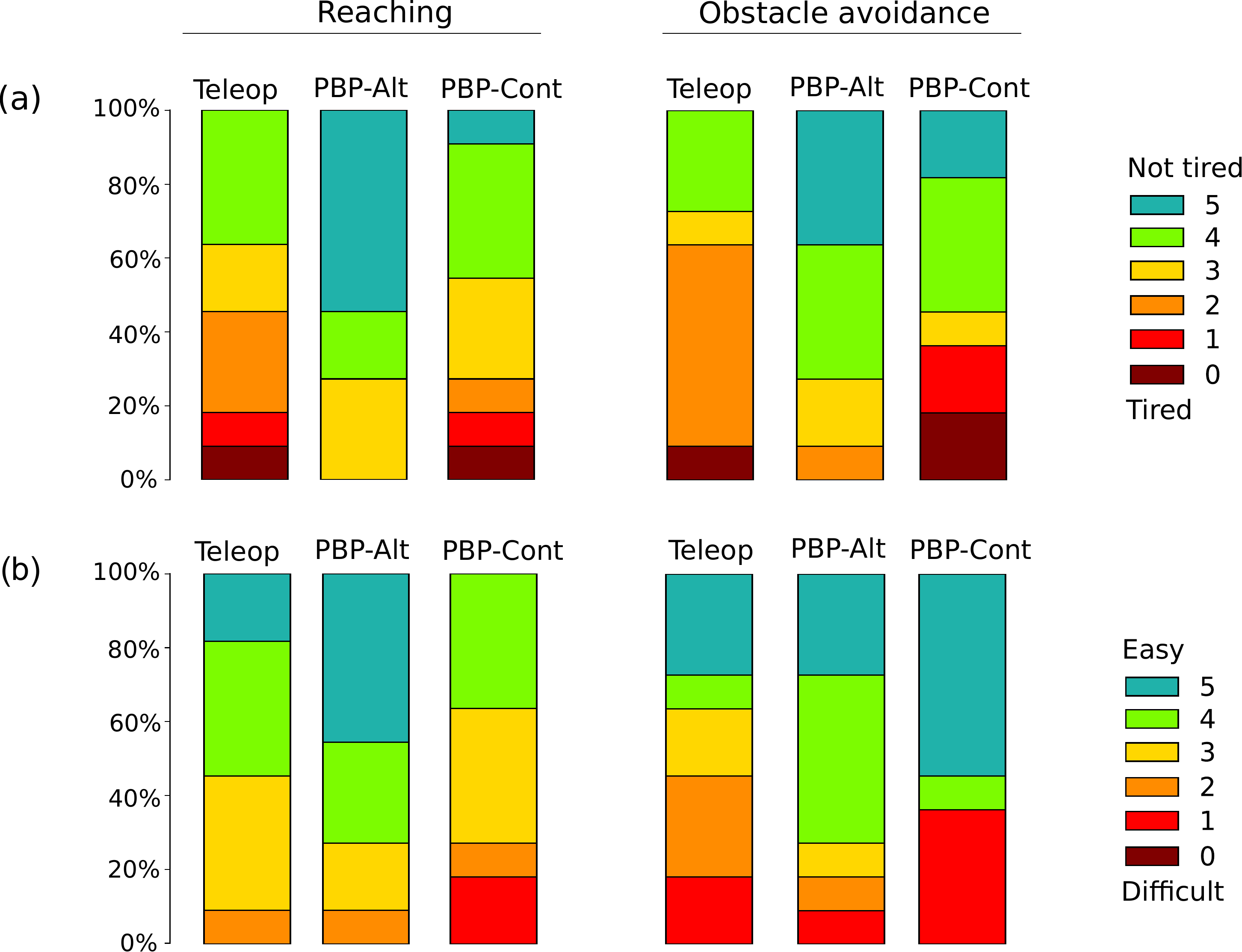}
	\vspace{-.2cm}
	\caption{
		The proportion of answers on the subjective evaluation of the three controllers on the reach and obstacle avoidance tasks concerning tiredness (a) and ease of operation (b).
		Overall, users gave a higher score to the PBP-Alt method while PBP-Cont showed the highest disparity. 
	}
	\vspace{-.2cm}
	\label{fig:userstudy}
\end{figure}

Fig. \ref{fig:userstudy} shows the subjective scores of the 11 participants.
They were asked to answer two questions: ``how tired did you feel?'' and ``how easy it was to operate the robot?''. The answer options were graded between 0 (bad) to 5 (good). To make relative ratings, the users answered only after using the three controllers. PBP-Alt received the largest proportion of the better scores, followed by Teleop and PBP-Cont. The Teleop mode presented the worst median value in terms of tiredness for both reaching and obstacle avoidance tasks. Statistical significance assessed with pairwise Wilcoxon signed-rank test with Bonferroni correction indicated that users considered the PBP-Alt less tiring to use than Teleop in both reaching and obstacle avoidance tasks ($p = 0.0446$ and $p=0.0485$, respectively).
Teleop and PBP-Alt were both considered equally easy to use with no statistically significant difference in both reaching and obstacle avoidance tasks ($p=0.7746$, $p=0.5566$, respectively). These results suggest that PBP-Alt is as easy to use as Teleop while being less tiring. No statistically significant differences were found between PBP-Cont and Teleop in regards to tiredness and ease of use in either task.

PBP-Cont received highly different scores in obstacle avoidance with approximate equal fractions for the worst and the best scores. Without Bonferroni correction, participants considered PBP-Alt easier to use than  PBP-Cont in the reaching task ($p=0.0419$). 
The scores suggest that the PBP-Alt provides a more preferable balance between receiving assistance without the impression of loss of control.
The disparity in user preference between PBP-Alt and PBP-Cont did not translate as differences in performance; similar results in terms of user intervention, task time, and collision ratios were observed in both cases (refer back to Fig. \ref{fig:user_metrics_and_collision_ICRA}).

\section{Conclusion}

By leveraging the inherent capability of primitives in assimilating external disturbances we presented PBP, a practical and simple method for shared control and assisted teleoperation. The method does not require an arbitration parameter and is capable of addressing multiple goals.
The user study showed that shared control with the proposed method led to comparable performance to teleoperation while significantly decreasing operator load. Users could effectively control the robot and indicate their intention towards different goals, and also decrease collisions with moving obstacles by re-shaping the primitives via teleoperation.
Operating the robot with PBP was no more difficult than teleoperation while being less tiring.
We are currently implementing PBP-Alt on the real HSR robot while controlling the 6-DoFs of the end-effector using a 3D mouse.
% (end of the video in: {\small \verb|https://youtu.be/rqq46_-Bwes|}).

\section{Acknowledgment}

The authors cordially express their gratitude to Mr. Shimpei Masuda and Dr. Koji Terada in Preferred Networks, Inc. and Mr. Koichi Ikeda, Mr. Hiroshi Bito, and Dr. Hideki Kajima in Toyota Motor Corporation for their assistance.

%\bibliographystyle{IEEEtran}
%\bibliography{teleop_with_primitives}


\end{document}